\documentclass[letterpaper, 10 pt, conference]{ieeeconf}

\IEEEoverridecommandlockouts

\usepackage{graphics}
\usepackage{amsmath}
\usepackage{amssymb} 
\usepackage{xcolor}
\usepackage[hidelinks]{hyperref}
\usepackage{graphicx}
\usepackage{wasysym}
\usepackage{ulem} 
\usepackage{tcolorbox}
\usepackage{booktabs}
\usepackage{siunitx}
\usepackage{textcomp}

\input{paper_style.sty}

\title{\LARGE \bf
Open Continuum Robotics -- One Actuation Module to Create them All
}

\author{
Reinhard M.~Grassmann, \textit{Student Member, IEEE,} 
Chengnan Shentu, \textit{Student Member, IEEE,} \\
Taqi Hamoda,
Puspita Triana Dewi,    
and %
Jessica Burgner-Kahrs, \textit{Senior Member, IEEE}%
\thanks{
We acknowledge the support of the Natural Sciences and Engineering Research Council of Canada (NSERC), [RGPIN-2019-04846] as well as the Canada Foundation for Innovation and Ontario Research Fund [Project \#40110].
}
\thanks{All authors are with Continuum Robotics Laboratory, Department of Mathematical and Computational Sciences, University of Toronto, Mississauga, ON L5L 1C6, Canada {\tt\small reinhard.grassmann@utoronto.ca}}%
\thanks{$^\dagger$OpenCR Project: {\tt\small \href{http://www.opencontinuumrobotics.com/}{www.opencontinuumrobotics.com}}}
}

\begin{document}

\maketitle
\thispagestyle{empty}
\pagestyle{empty}

\begin{abstract}
Experiments on physical continuum robot are the gold standard for evaluations.
Currently, as no commercial continuum robot platform is available, a large variety of early-stage prototypes exists.
These prototypes are developed by individual research groups and are often used for a single publication.
Thus, a significant amount of time is devoted to creating proprietary hardware and software hindering the development of a common platform, and shifting away scarce time and efforts from the main research challenges.

We address this problem by proposing an open-source actuation module, which can be used to build different types of continuum robots.
It consists of a high-torque brushless electric motor, a high resolution optical encoder, and a low-gear-ratio transmission.
For this letter, we create three different types of continuum robots. 
In addition, we illustrate, for the first time, that continuum robots built with our actuation module can proprioceptively detect external forces.
Consequently, our approach opens untapped and under-investigated research directions related to the dynamics and advanced control of continuum robots, where sensing the generalized flow and effort is mandatory.
Besides that, we democratize continuum robots research by providing open-source software and hardware with our initiative called the Open Continuum Robotics Project$^\dagger$, to increase the accessibility and reproducibility of advanced methods.
\end{abstract}

\copyrightnotice

\section{INTRODUCTION}

Research in the field of continuum robotics is characterised by custom-built prototypes harming the research community.
Proprietary prototypes are designed and manufactured in their respective labs, and operated with proprietary software.
Alarmingly, our study \cite{GrassmannBurgner-Kahrs_et_al_RSS_WS_2020} on a specific continuum robot type reveals that \SI{61.1}{\%} of all prototypes are used for only a single publication.
This trend can be observed for other continuum robot types as well.
The large and growing number of different prototypes imposes a barrier for the research community as time and focus are directed to building custom prototypes which hinders the development of advanced hardware and software \cite{GrassmannBurgner-Kahrs_et_al_RSS_WS_2020}.
To mitigate these drawbacks, one might suggest to opt for an open-source approach.

Probably due to most prototypes' short lifespan of one publication, position controlled servo motors are used, thus lacking capabilities of measuring and controlling of the flow and effort of the system, e.g., tendon tension or actuation force.
This limitation further reduces the applicability of inverse dynamic models along with their advantages.
For example, impedance control %
or potential field %
methods cannot be properly utilized.
We believe that providing torque-controlled motors facilitates research in advanced control and learning algorithms.

\begin{figure}
    \centering
    \vspace{0.75em}
    \includegraphics[width=0.90\columnwidth]{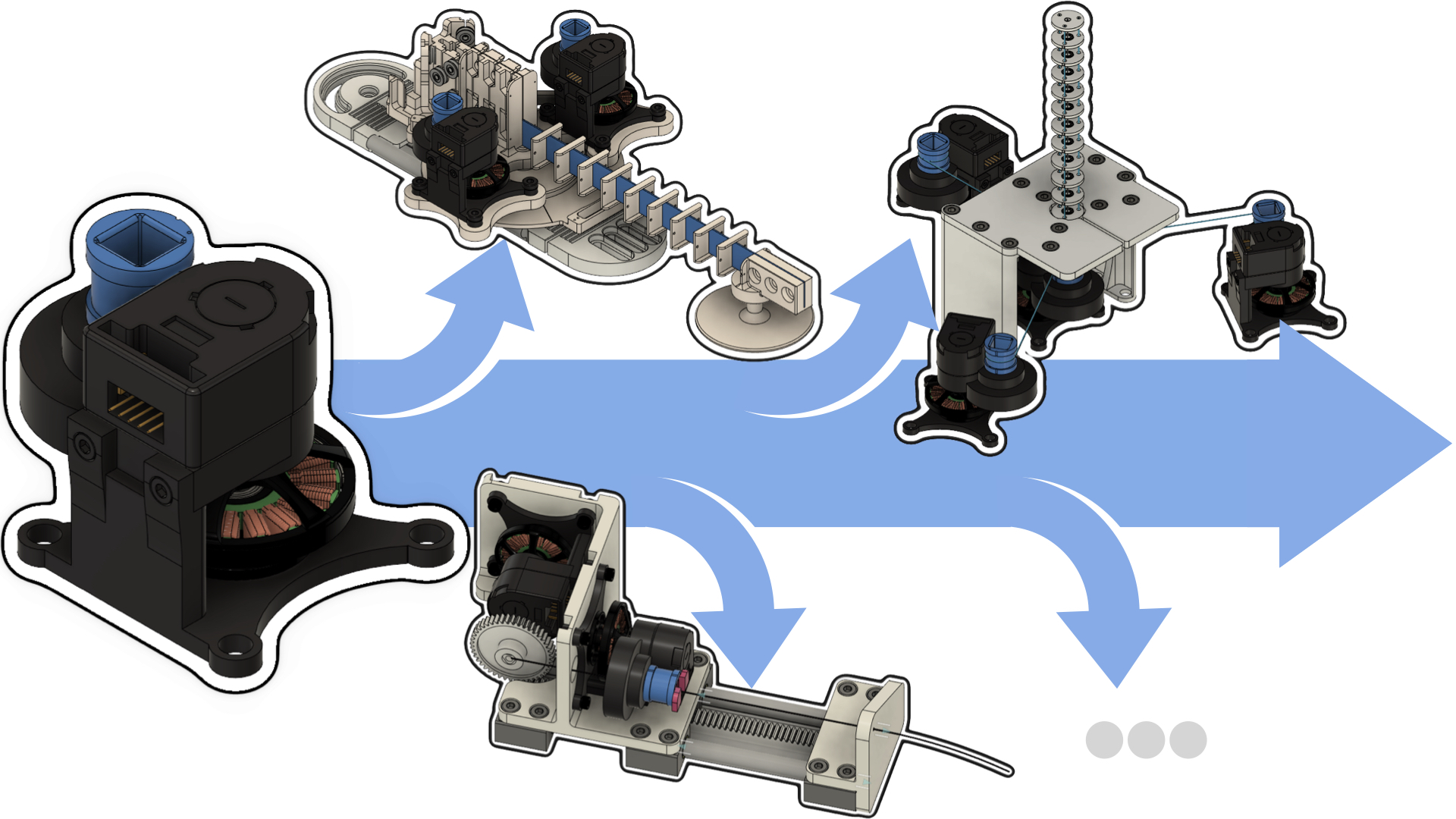}
    \caption{
    An actuation module applicable to generate tendon displacement/tension as well as for backbone rotation/torque for a wide range of continuum robots.
    }
    \label{fig:catchy_image}
\end{figure}

To take a step in the direction towards torque-controlled continuum robots, we introduce an actuation module for continuum robots suitable for linear translation, axial rotation, and tendon-introduced motion.
For the sake of clarification, these types of motion are of interest for small-diameter continuum robots that are either actuated by tendons or realized as concentric tube arrangements.

With this letter, we provide the CAD designs and assembly instructions for the actuation module alongside an interface for position and torque control.
Using this actuation module, we present three example continuum robot prototypes showing the versatility and modularity: a planar and a spatial tendon-driven continuum robot (TDCR) and a concentric tube continuum robot (CTCR) as depicted in Fig.~\ref{fig:catchy_image}.
All designs, component lists, assembly instructions, and software are open-source and provided to researchers and practitioners on \href{http://www.opencontinuumrobotics.com/}{www.opencontinuumrobotics.com}.

\section{RELATED WORK}
\label{sec:related_work}

In this section, we discuss the current state of prototyping in our research community and highlight three missing characteristics, which we argue to be beneficial for future prototypes: (i) the capability of performing high frequency control and advanced torque control, (ii) modularity to unify different continuum robot types, and (iii) accessibility of the hardware and software.

\subsection{Continuum Robot Hardware}

In continuum robotics, DC motors with high gear-ratio are widely used to generate angular motion, see \cite{NwaforRabenorosoa_T-RO_2023}.
For linear motion, DC motors with lead screws and nuts are used, e.g., \cite{XuSimaan_T-Ro_2008, AmanovNguyenBurgner-Kahrs_IJRR_2021,Sarma2022}.
Both approaches for generating motion prevent the proprioceptive measurement of the effort, due to the lack of transparency \cite{WuethrichBauer_et_al_CoRL_2020} and backdrivability \cite{IshidaTakanishi_RAMech_2006}.
Currently, to measure effort, authors use force sensors or load cells, e.g., \cite{XuSimaan_T-Ro_2008, AloiRucker_et_al_RA-L_2022, DeutschmannReineckeDietrich_RoboSoft_2022}.
However, using additional sensors adds cost, computational load, and weight.
Furthermore, they may be too large for small-scale continuum robots.
In contrast, measuring the motor current comes with very low additional burden and can be accomplished in a high reading rate, since motor current is part of the low-level motor control.

\subsection{Torque Controlled Robot}

Continuum robot control methods mostly address end-effector pose while assuming quasi-static motion in free or static environments.
Although satisfactory results have been achieved, control of more advanced states -- such as the robot's shape, distributed forces and stiffness, and kinematic performances -- remains largely unsolved.
Moreover, results presented are not generalizable to other continuum robot types.
To push forward the current applications, multiple control requirements must be met.
Besides task space requirements such as accuracy and robustness to disturbances, authors in \cite{ChikhaouiBurgner-Kahrs_ACTUATOR_2018} point at real-time state-estimation to enable active compliance and force applications.

Looking at other robotic research fields, legged robots face the same set of requirements when performing dynamic locomotion in unstructured environments, and they utilize torque control to meet these requirements \cite{WensingKim_et_al_T-RO_2017, GrimmingerRighetti_et_al_RA-L_2020}.
Torque control has a rich history, especially for serial kinematic mechanisms \cite{LuhFisherPaul_TAC_1983, LawrenceKhatibHake_TRA_1989, BischoffHirzinger_et_al_ISR_2010, HaddadinHaddadin_et_al_RAM_2022}. %
The spread and success of torque controlled robots in these area of robotics has not translated into continuum robots - probably for the lack of a common robot platform with torque control capabilities.

\subsection{Modular Robots}

It is important to design prototypes with reliability, low maintenance effort, and low cost in mind.
An effective strategy to achieve these criteria is to design for modularity by reducing the number of different components and lowering the cost and difficulty of maintenance, which in return keeps the system reliable and scalable.
Modularity also helps build complex robots with higher degrees of freedom by extending the initial work.
For instance, this approach can be seen in hyper-redundant snake robots \cite{WrightHowie_IROS_2007}, legged robots \cite{SeokKim_et_al_IROS_2012, WensingKim_et_al_T-RO_2017}, articulated arms \cite{BischoffHirzinger_et_al_ISR_2010, HaddadinHaddadin_et_al_RAM_2022}, and humanoids \cite{ChignoliKim_et_al_Humanoids_2020}.
Furthermore, modularity allows to bridge different domains in robotics, where modular components are developed for one domain, e.g., \cite{GrimmingerRighetti_et_al_RA-L_2020, SeokKim_et_al_IROS_2012}, and can be reused or adapted in another domain, e.g., \cite{WuethrichBauer_et_al_CoRL_2020, ChignoliKim_et_al_Humanoids_2020}.
To the best of our knowledge, there is no previous work in designing a modular actuator for continuum robots.
Existing prototypes use off-the-shelf motors but are composed of custom linking or transmission mechanisms.

\subsection{Open-Source Robotics and OpenCR Project}

Open-source robotics benefit the research community, where commercial robotics hardware is costly or non-existent.
With recent progress in rapid prototyping, open-source robotics provide the user with the flexibility to modify designs or explore alternative components to meet their applications.
This approach has gained success in research areas such as humanoid robotics \cite{MettaNori_et_al_PerMIS_2008}, surgical robotics \cite{HannafordWhite_et_al_TBME_2012}, soft robotics \cite{HollandWalsh_et_al_SR_2014}, robotic learning \cite{AhnLevine_et_al_CoRL_2020, WuethrichBauer_et_al_CoRL_2020}, and legged locomotion \cite{GrimmingerRighetti_et_al_RA-L_2020}.
An in-depth survey of more than \num{80} major projects and initiatives across different robotic communities is provided by \cite{PatelLiarokapisDollar_RAM_2022} highlighting the clear benefits and associated challenges.
In our research community, continuum robotics, an incipient trend towards releasing software \cite{TillAloiRucker_IJRR_2019, SadatiWalker_et_al_IRJJ_2021, RaoBurgner-Kahrs_et_al_Frontiers_2021, Janabi-SharifiJalaliWalker_Access_2021, BentleyRuckerKuntz_Access_2022}, hardware \cite{BlumenscheinHawkes_et_al_Frontiers_2020, ClarkMathivannanRojas_TMRB_2021, DeutschmannReineckeDietrich_RoboSoft_2022}, tutorials \cite{TillAloiRucker_IJRR_2019, RaoBurgner-Kahrs_et_al_Frontiers_2021, Janabi-SharifiJalaliWalker_Access_2021} and datasets \cite{GrassmannBurgner-Kahrs_et_al_IROS_2022} can be observed.
While isolated small projects exist, an encompassing project promoting research in the same way as described in \cite{PatelLiarokapisDollar_RAM_2022} is sorely needed yet still missing.

To date, there are a multitude of barriers that continuum roboticists must overcome if they want to get into continuum robotics, including but not limited to: no textbook on this topic, research paper paywalls, limitations in robotic simulator capabilities, and a lack of obtainable continuum robotics hardware.
In the interest of promoting increased exploration in the field and, thus, new scientific discoveries, we launch the \href{http://www.opencontinuumrobotics.com}{Open Continuum Robotics Project (OpenCR Project$\dagger$)} to release software code and hardware-related files for the purpose of reproducibility, verification, and benchmarking.

\section{ACTUATOR DESIGN AND REALIZATION}

\begin{figure*}[!h]
    \centering
    \vspace{0.75em}
    \includegraphics[width=0.825\textwidth]{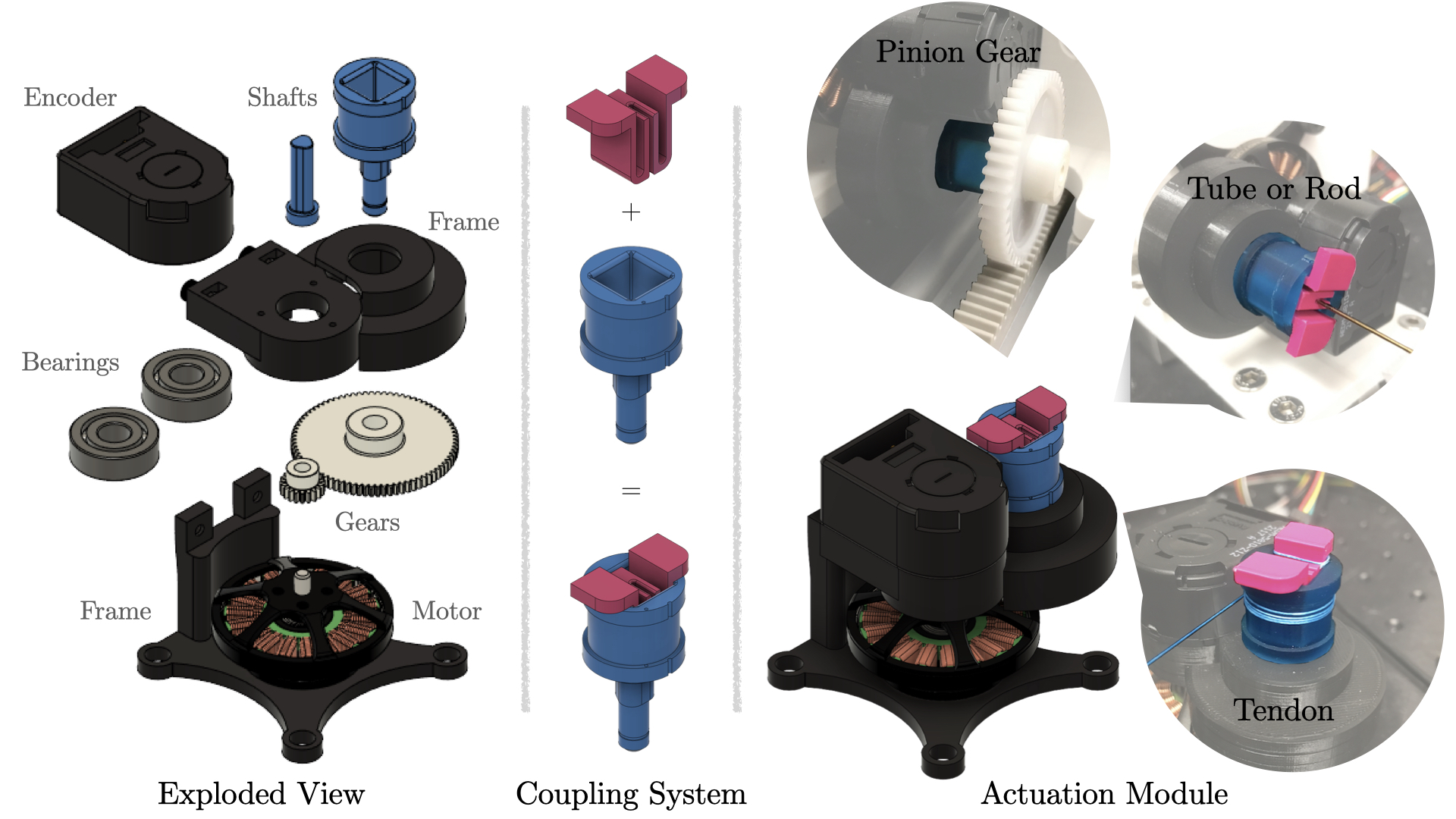}
    \caption{
    Actuation module with coupling system. 
    (left) Exploded view of a actuation module. 
    (middle) Coupling system, where the second part can be changed on the fly.
    (right) Actuation module with different use cases for specific motion modes. 
    To pull on a tendon, the tendon is terminated and spooled on the drum. For translational motion, a pinion gear is inserted. 
    To rotate a backbone or tube, the tube or rod is attached to the coupling system, which has a through hole.
    }
    \label{fig:hardware}
\end{figure*}

In this section, the actuation module and our design choices are described.
All details necessary for building the proposed actuation module as well as the derived prototypes are open source on the \href{http://www.opencontinuumrobotics.com/}{OpenCR Project website$^\dagger$}.

\subsection{Requirements and Properties}

The goal is to design a compact and versatile actuation module capable of measuring and controlling velocity and force related quantities such as tendon displacement and tendon tension. 
To increase versatility, a compact form factor that allows for usage in different prototypes is desirable similar to the design proposed in \cite{MengacciBicchi_et_al_TMECH_2021} for humanoid robots, serial kinematic arms, and more. 
To utilize advanced control approaches, the actuation module should be embedded in a real-time system.
A high-level controller should run at least with \SI{1}{kHz}, while the low-level controller should run at least one magnitude faster than the high level controller. 
Furthermore, advanced control approaches should be able to have access to general flow and effort \cite{GawthropBevan_MCS_2007} being velocity and force related quantities. 
To measure effort, i.e., external torque in the joint level, the actuation module can either be equipped with a torque sensor or can be designed in a way such that proprioception is possible. 
We opt for the latter to reduce cost.
Therefore, an actuation module needs to have low transparency and should be backdrivable as mentioned in \cite{SeokKim_et_al_IROS_2012} to allow for proprioceptive measurement via the motor current.
This design choice requires an actuator with high torque density \cite{SeokKim_et_al_IROS_2012, WensingKim_et_al_T-RO_2017}, which also enables fast motion.
As used in \cite{GrimmingerRighetti_et_al_RA-L_2020, WuethrichBauer_et_al_CoRL_2020}, brushless electric motors used for aerial drones provide an affordable alternative to custom motors with high torque density.
Brushless electric motors also reduce the noise in the motor current, which is essential for measuring effort in a proprioceptive manner.
Moreover, it should be noted that a small tendon displacement can lead to high curvature and bending of a segment \cite{Grassmann_OpenCR_2023}. 
Especially, when using a small gear ratio, the motor angle will be small requiring the measurement small motor angle displacements with high encoder resolution.

Our proposed continuum robot actuation module has the following inherent properties:
\begin{itemize}
    \item High-performance brushless electric motor providing high-torque actuation.
    \item A combination of a brushless electric motor and a low-gear-ratio transmission enabling proprioceptive torque sensing by measuring motor currents.
    \item \SI{10}{kHz} low-level controller running on a micro-controller to enable fast motor control.
    \item \SI{1}{kHz} high-level controller running on a consumer computer with a real-time patched Ubuntu operating system.
    \item Ability to sense external forces and react to them in real time.
    \item A fairly simple design consisting of off-the-shelf and custom 3D printed parts, lowering the total cost.
\end{itemize}
For the mechanics and electronics of the proposed actuation module, we took inspiration from the \href{https://open-dynamic-robot-initiative.github.io/}{Open Dynamic Robot Initiative}, where efforts are made to build cost efficient actuators for torque controlled legged robots and manipulators, e.g., \cite{GrimmingerRighetti_et_al_RA-L_2020, WuethrichBauer_et_al_CoRL_2020}.
As we describe in the following, we use similar hardware and electronics components when possible to leverage robotics community knowledge and support knowledge mobilization.

\subsection{Motor, Gearbox, and Mechanical Frame}

The actuation module shown in Fig.~\ref{fig:hardware} is a low-cost and low-complexity design. 
It consists of off-the-shelf parts and custom 3D printed parts.
The off-the-shelf parts include a brushless electric motor (Antigravity MN4004-25 KV300, T-motor, Jiangxi, China), a pair of molded acetal gears (0.5 Module, 80 teeth and 20 teeth) forming a 4:1 transmission, a high-resolution optical encoder (AEDM-5810-Z12, Avago, CA, US) with a \num{5000} counts-per-revolution code wheel, and two steel ball bearings (inner diameter of \SI{8}{mm} and outer diameter of \SI{22}{mm}).
The motor shaft, joint shaft, and the frame are custom 3D printed parts.
The shafts are printed with a resin 3D printer, while a 3D printer with filament is used to print the frame.
We refrain from custom made machined parts due to the cost of production for small lot sizes.

The assembly of the actuation module is straight forward.
It requires two additional M3 screws and nuts, which hold the two-part frame together.
The code wheel of the optical encoder is mounted directly on the motor shaft. 
The frame provides two slots to integrate the ball bearings.

To ensure sufficient transparency, the transmission ratio is low, i.e., 4:1.
Note that the term transparency is closely related to the concept of backdrivability of
an actuator \cite{SeokKim_et_al_IROS_2012}.
In our case, transparency means that the torque applied at the joint shaft directly translates to torque at motor shaft and, therefore, is translated to the motor itself without being absorbed by a high gear-ratio transmission, friction between gears, and high mechanical impedance.
This enables accurate torque control and torque measurements through motor current measurements avoiding additional sensors such as tendon-tension or force and torque sensors.
Furthermore, a low transmission ratio enables a reasonable peak torque and high velocity at the joint.
In contrast, common continuum robots as described in Sec.~\ref{sec:related_work} rely on actuators with a large gear-ratio leading to a non-backdrivable actuator without transparency.

\subsection{Joint Shaft with Coupling Device}

To perform different modes of motion, such as tube rotation and translation for CTCR or tendon displacement and tendon tension for a TDCR, we propose a coupling device.
Figure~\ref{fig:hardware} illustrates the coupling device and its utilization for a selection of different actuation modes.
The coupling device consists of two parts.
The first part is designed as a hollow shaft and has a square hole.
The shaft is rigidly attached to a gear and aligned via a bearing.
It is hollow to allow tubes and rods to be inserted.
For winding a tendon, it also features a drum with a radius of \SI{9}{mm} acting as a winch.
The second part of the coupling device is used to attach additional mechanisms such as a tube holder or a pinion gear, see Fig.~\ref{fig:hardware}.

The coupling system is simple yet efficient to attach transmission mechanism for -- but not limited to -- tubes, tendons, or pinion gears.
Therefore, the actuation module is robot-agnostic and new prototypes can be realized easily. 
It can be used independently of the prototypes proposed in the present letter.

\subsection{Electronics and Communication}
\label{sec:electronics_and_communication}

The electrical components are similar to the ones used by Open Dynamic Robot Initiative described in \cite{WuethrichBauer_et_al_CoRL_2020}.
We use an off-the-shelf evaluation board (LAUNCHXL-F28069M, Texas Instruments, Dallas, USA) with a micro-controller.
To enable field oriented control for the brushless electric motor in the actuation module, an off-the-shelf booster card (BOOSTXL-DRV8305EVM, Texas Instruments, Dallas, USA) is utilized.
Note that one micro-controller can be equipped with two booster cards and, further, the custom electronic boards designed in \cite{GrimmingerRighetti_et_al_RA-L_2020} can be used to reduce the cost even further.

The electronics and communication are chosen to be fast.
While a low-level controller can be implemented on the evaluation board, a PC is used to run a high-level controller.
A real-time capable PC running Ubuntu patched with RT-Preempt is connected to the evaluation board.
Control frequencies of \SI{10}{kHz} and \SI{1}{kHz} for low-level and high-level controllers, respectively, are possible.
A CAN-to-PCI Express interface card (IPEH-003027, PEAK system, Hessen, Germany) is used for communication between the PC and the evaluation board.

\section{ACTUATOR EVALUATION}

In this section, we evaluate the performance of the proposed actuation module under static and dynamic load.
The capabilities and characteristics are illustrated by different assessments on two different toy examples.
For the sake of brevity, we omit details on the toy examples which are provided on our website$^\dagger$.
Without lost of generality to other possible actuation modes, we limit the assessments to tendon displacement and tendon tension.
In this case, a positive tendon length corresponds to the amount of tendon pulled by the actuation module.
Moreover, we omit the plots for motor current, and focus on the utilization for continuum robots, since linear relation between torque and current is well established in the literature.

\subsection{Empirical Gain Tuning}

\begin{figure}
    \centering
    \vspace{0.75em}
    \includegraphics[width=0.95\columnwidth]{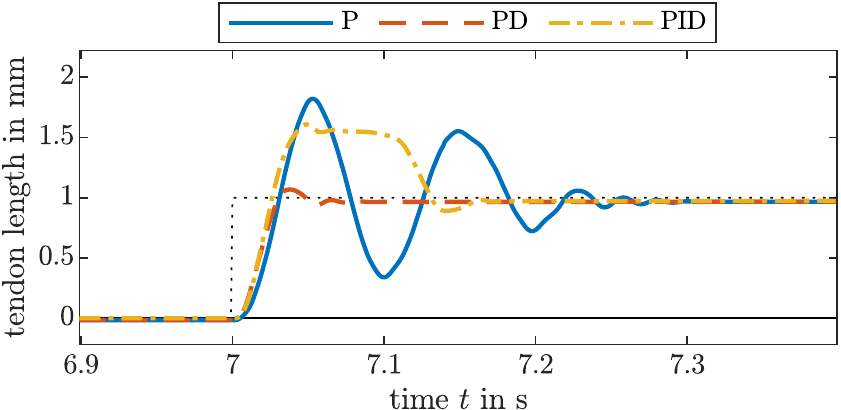}
    \caption{
    Step responses of the actuation module with controllers tuned by the standard Ziegler-Nichols method.
    The controller gain $k_\text{P}$ is \SI{1.0}{} for the P controller.
    The PD controller gains are $k_\text{P} = 1.6$ and $k_\text{D} = 0.02$, whereas the gains for the PID controller are $k_\text{P} = 0.9$, $k_\text{I} = 10.1$, and $k_\text{D} = 0.02$.
    }
    \label{fig:toyexample_zieglerNichols}
\end{figure}

We used the Ziegler-Nichols method to find sets of controller gains. 
The step input is set to \SI{1}{mm} assuming a tendon actuation.
This corresponds to about \num{353} counts on the code wheel.
Note that tendon displacement is proportional to the motor angle.
Figure~\ref{fig:toyexample_zieglerNichols} shows the step response of a P, PD, and PID controller.
Evaluated on the real-time PC, a motor position control outputs the motor current, which will be sent to the evaluation board at a frequency of \SI{1}{kHz}.
Throughout the subsequent assessments, we use the same controller gains for all motors. 
To further improve quantitative results, fine tuning of each constant controller gain for each actuation module is advisable as the continuum robot structure will significantly influence the performance.
Therefore, utilizing model-based controller gains is the way to go and subject of future work.
Note that the Ziegler-Nichols method results in an aggressive controller, e.g., large overshoot as can be seen in Fig.~\ref{fig:toyexample_zieglerNichols}.

Additionally, we tested larger and smaller step inputs.
On the one hand, a large step input of \SI{10}{mm} results in a set of controller gains being insensitive to small changes, i.e., below \SI{2}{mm}.
This is undesirable for short distances and slow trajectories, where the differences in tendon length between adjacent time steps are too small.
On the other hand, ultimate gains of Ziegler-Nichols method derived from small step input of \SI{0.1}{mm} was too large reacting to small disturbances and causing chattering behaviour and, ultimately, unstable behaviour.
Note that the electric motor used are designed for drones, thus has lower inertia and better transparency compared to other electric motors, with less mechanical damping as the trade-off.

\subsection{Static Load Compensation via Tendon Tension Offset}

The purpose of this evaluation is to demonstrate the actuator's feasibility for fast joint motion under load.
Different static loads are applied by hanging weights (\SI{200}{g}, \SI{540}{g}, and \SI{1000}{g}) on the tendon.
The tendon is pulled by the actuation module and redirected by a pulley such that the tendon tension is equal to the gravity pull on the weight.

We used two controllers, namely PD and PD-$g(\theta)$ with the tuned gains described above -- a PD-$g(\theta)$ is a PD controller with gravity compensation.
Since the gravitational pull is constant with respect to the tendon displacement, the external load appears as a constant offset in the motor current.
The compensation in PD-$g(\theta)$ is fairly simple, since the weight is proportional to the motor current and invariant with respect to the motor angle $\theta$.
The load of \SI{200}{g}, \SI{540}{g}, and \SI{1000}{g} are considered by adding a constant current of \SI{0.2}{A}, \SI{0.54}{A}, and \SI{1}{A}, respectively, to the controller output.

\begin{figure}[h]
    \centering
    \includegraphics[width=0.95\columnwidth]{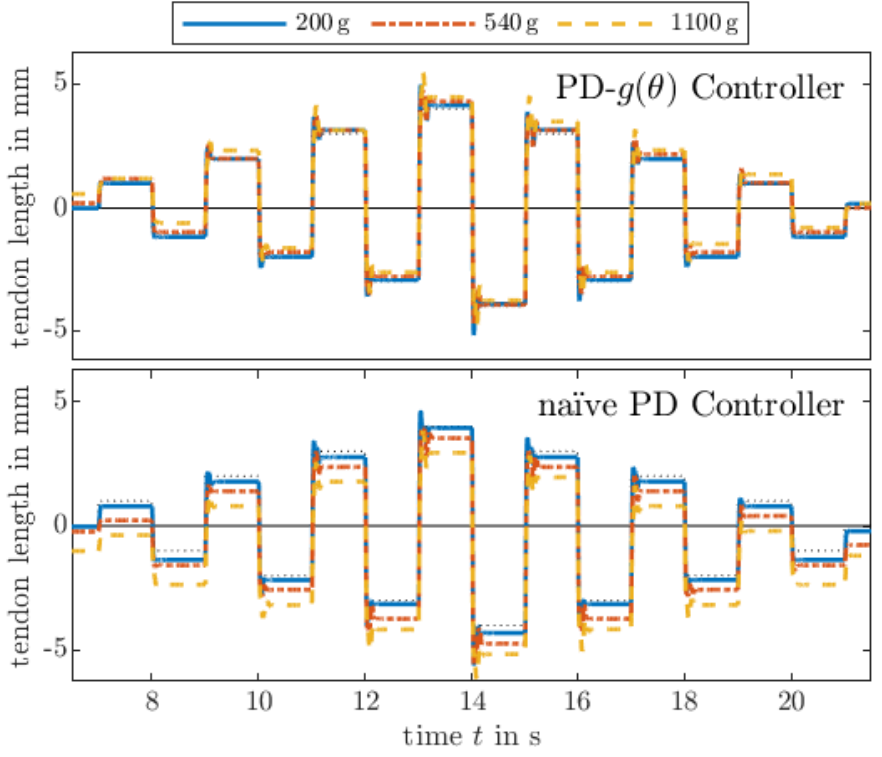}
    \caption{
    Comparison of PD controllers with and without gravity compensation.
    The desired step is shown as a black dashed line. 
    It is evident from the step responses that the higher the load, the higher the steady-state error of the na\"ive PD controller. 
    Furthermore, a qualitative comparison shows a predictable pattern -- each steady-state error is (nearly) proportional to the load, which can be compensated by adding a constant current to the na\"ive PD controller leading to a PD-$g(\theta)$ controller.
    Step responses of different heights with different tendon tensions show that the PD-$g(\theta)$ controller successfully compensates the known load, while the steady-state errors of a na\"ive PD controller are relative to the load.
    }
    \label{fig:toyexample_weights_plots}
\end{figure}

Figure~\ref{fig:toyexample_weights_plots} shows step responses for different alternating heights, i.e., between \SI{-8}{mm} and \SI{7}{mm} in \SI{1}{mm} steps.
The results show a reasonable response.
As expected, it can be observed that the PD-$g(\theta)$ controller generally performs better than the na\"ive PD controller in terms of steady-state error.
Potential cases for the remaining error include the steady-state error of a PD controller, friction between pulley and tendon, as well as friction between axle and the axle holder of the pulley.
Note that the tendon moves relatively fast and the pulley has an inertia and internal friction, hence the friction between the tendon and the pulley.

\begin{figure}[h]
    \centering
    \includegraphics[width=0.95\columnwidth]{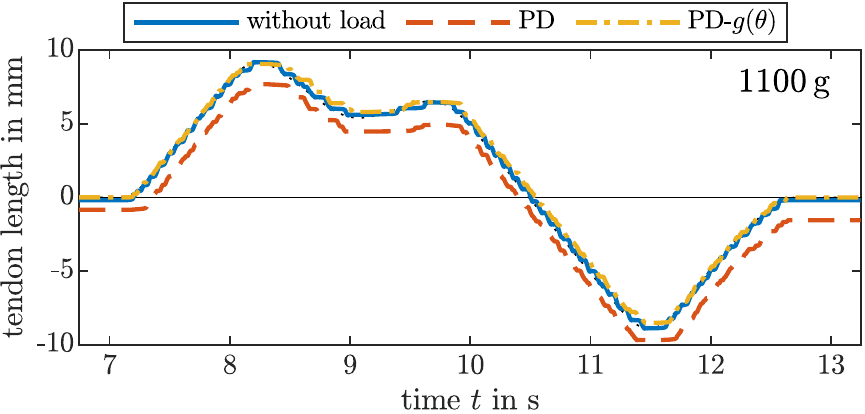}
    \caption{
    Following a desired $\mathcal{C}^4$ smooth trajectory with and without load and gravity compensation.
    The uncompensated load appears as a constant offset in the error.
    The performance of the PD-$g(\theta)$ controller with load coincides whit the performance of the PD controller without load.
    Note that, in an unloaded case, both controller behave similarly.
    }
    \label{fig:toyexample_weights_trajectory}
\end{figure}

Furthermore, the controllers are tested on a smooth trajectory.
We use the approach described in \cite{GrassmannBurgner-Kahrs_RA-L_2019} to generate a $\mathcal{C}^4$ smooth trajectory including position and velocity for the motor angle and tendon length, respectively.
Figure~\ref{fig:toyexample_weights_trajectory} shows the response for the unloaded and loaded cases.
For the latter, the controller either does or does not compensate the external load of \SI{1100}{g}.
Note that the controllers are simple error-driven controllers meaning no proper trajectory tracking is used.
Nevertheless, a similar conclusion from the step response assessment can be made:
Gravity compensation via PD-$g(\theta)$ can account for the external load by adding a constant offset of \SI{1.1}{A}.

For short distances and slow velocities, tendon displacement is jerky.
While the jittering is in the sub-millimetre range, it causes high changes in the velocity profile.
This effect almost vanishes for very high velocities.
However, changes in tendon length are rather small and, therefore, slower velocities are necessary.
In addition to the error-driven nature of the used controllers, the brushless electric motor inherently suffers from ripple effects due to cogging torque generated by the permanent magnets.
A controller with a feedforward term to counteract the ripple effects might reduce the disturbance.

\subsection{Variable Tendon Tension due to Beam Flexibility}

In general, a continuum robot has a flexible backbone which can store elastic energy.
To show that the proposed actuation module can proprioceptively measure the effect of the stored elastic energy, we conduct two assessments using a toy example.
This toy example is composed of one actuation module and a flexible \SI{285}{mm} flat beam as a backbone.
One tendon is guided by spacer disks along the backbone.
By pulling on the tendon and applying tendon tensions, the backbone can be deformed.
The deformation happens in a bending plane which is orthogonal to the direction of the gravity.
Therefore, the to be observed effect is due to the beam's flexibility whereas the gravitational load can be neglected.
While a gravitational load appears as constant tendon tension, constant error, and constant motor current as shown above, the effect here should appear as variable error depending on the tendon displacement.

\begin{figure}[b]
    \centering
    \includegraphics[width=0.95\columnwidth]{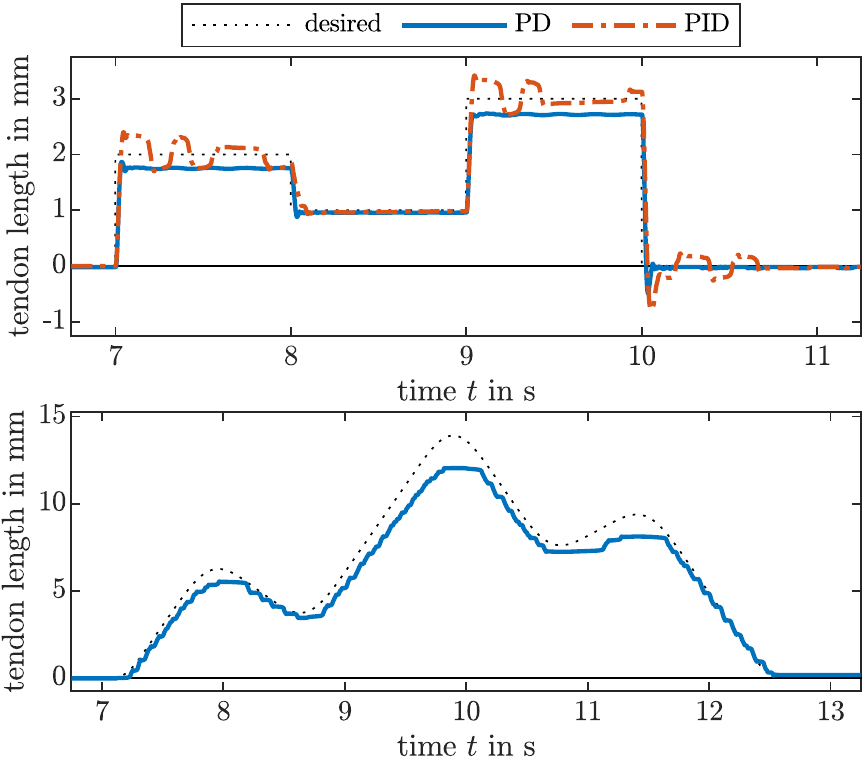}
    \caption{
    An actuated flexible beam response to step inputs and a desired $\mathcal{C}^4$ smooth trajectory.
    The flexible beam induces a variable tendon tension appearing as a variable steady-state and trajectory tracking error.
    The error correlates with the amount of bending.
    Note that we omit the PID controller for the lower panel to highlight the bending-variant tendon tension. 
    A constant and fine-tuned integrator gain could compensate for the error.
    }
    \label{fig:toyexample_beam_plots}
\end{figure}

For the first assessment, step responses using the same PD and PID controller from above are recorded for the toy example, see Fig.~\ref{fig:toyexample_beam_plots}.
The steady-state error of the PD controller for step height \SI{2}{mm} and \SI{3}{mm} are similar.
This is expected because small changes in tendon length will not result in a high bending and, therefore, a high energy storage.
The PID controller can reduce the error but starts to oscillate indicating that a na\"ive implementation of the integrator is not sufficient.
A good set of fine-tuned PID gains can further compensate for the steady-state error. 
However, the magnitude of the observed oscillation is small and might be linked to a ripple-effect due to cogging torque as observed in the previous static load evaluation. 
Both controllers overshoot at the end generating negative values for the tendon length.

\begin{figure*}[h]
    \centering
    \vspace{0.75em}
    \includegraphics[width=.925\textwidth]{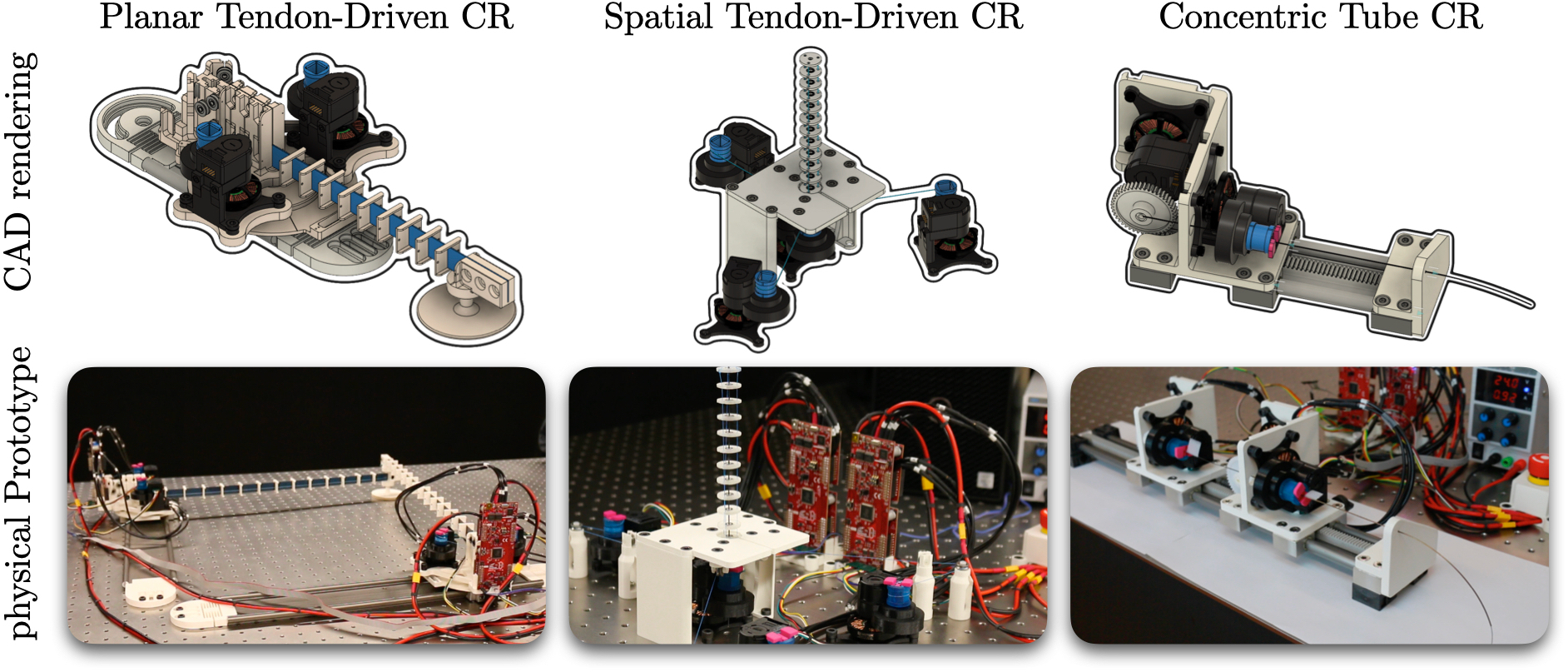}
    \caption{
    Renderings and built prototypes utilizing the proposed actuation module.
    }
    \label{fig:prototypes}
\end{figure*}

For the second assessment considering larger bending, a $\mathcal{C}^4$ smooth trajectory \cite{GrassmannBurgner-Kahrs_RA-L_2019} is executed by the toy example.
Figure~\ref{fig:toyexample_beam_plots} shows the desired and executed path.
It can be observed that the bending induces a higher error, which varies with the bending and relates to the spring-like behaviour of the flexible beam.
The higher the displacement, the higher the stored energy. 
This results in a variable error with respect to the tendon length, cf. Fig.~\ref{fig:toyexample_weights_trajectory}, where the error is constant.
Recall that the actuation module is backdrivable and the error can be compensated by offsetting the commanded motor current as shown for the first toy example.
Hence, the effect of the flexible beam acting as an additional tendon tension is visible as an error in the tendon displacement.
Furthermore, note that the actuation module is also transparent and the motor torque is proportional to the motor current.
Therefore, the actuation module can proprioceptively measure the variable load due to the flexible beam.

\section{BUILT CONTINUUM ROBOTS} %
\label{sec:prototypes}

Using the actuation module, one can build different continuum robots.
To highlight this, in this section, we briefly describe three different prototypes utilizing the proposed actuation module (see Fig.~\ref{fig:prototypes}).
All prototypes use a subset of possible motions -- tendon displacement, tube rotation, and carriage translation.
Further details, such as CAD drawings, components lists, and assembly instructions, are provided on the \href{http://www.opencontinuumrobotics.com/}{Open Continuum Robotics Project} webpage$^\dagger$.

\subsection{Planar Tendon-Driven Continuum Robot Prototype}

Similar to the second toy example, a planar TDCR is composed of a flexible flat beam as a backbone.
Spacer disks along the backbone guide the tendons.
By changing the tendons’ length and applying tendon tensions, the backbone can be deformed.
A pair of two actuation modules is used forming an antagonistic actuation pair.
We utilize two planar TDCRs to create a parallel continuum robot (PCR).
Both beams are linked together at the distal end via a passive revolute joint.
The proximal end of each beam is fixed to the respective base, while each base is linked to a platform via a passive revolute joint.
A platform can be manually adjusted and fixed along a rail.
Figure~\ref{fig:prototypes}(left) shows a rendering and the built prototype.

\subsection{Spatial Tendon-Driven Continuum Robot Prototype}

A spatial TDCR, shown in Fig.~\ref{fig:prototypes}(middle), is adapted from our previous work \cite{GrassmannBurgner-Kahrs_et_al_Frontiers_2022} and has three degrees of freedom in the task space. 
It has a center backbone made of a super-elastic Nitinol rod, and spacer disks along the backbone that guide the tendons. 
All spacer disks, except the one at the distal end of the segment, have a small ball-bearing placed between the backbone and the tendons, resulting in a passive rotational degree of freedom.
One actuation module at the center rotates the backbone and the distal spacer disk, actively changing the tendon routing \cite{GrassmannBurgner-Kahrs_et_al_Frontiers_2022}.
Spatial bending is achieved by three actuators.
In total, four actuation modules are used.

\subsection{Concentric Tube Continuum Robot Prototype}

A CTCR is composed of multiple nested tubes which are concentric, pre-curved, and super-elastic.
Each tube is typically attached to a carriage that can translate and rotate the tube relative to the other tubes.
We build two carriages with two actuation modules each, see Fig.~\ref{fig:prototypes}(right).
This results in a prototype with four degrees of freedom in joint space.

The translational motion of a carriage is realized with a rack-and-pinion mechanism which transforms rotational motion of the coupling mechanism to linear motion of the carriage.
Therefore, a pinion gear attached to the coupling mechanism of one of the actuation modules.
To complete the rack-and-pinion mechanism, a rack is rigidly attached to a rail, see Fig.~\ref{fig:hardware}.

\subsection{Financial Cost}

The estimated cost of each prototype is \SI{2127}{USD}, \SI{1682}{USD}, and \SI{1785}{USD}, for PCR, TDCR, and CTCR, respectively.
The estimations include the actuation modules, which cost \SI{212}{USD} per module, and the electrical components stated in Sec.~\ref{sec:electronics_and_communication} to run two actuation modules, which cost around \SI{240}{USD}.
To estimate the costs, we summarize the costs for the mechanical parts and add the costs for 3D printed parts by consulting different 3D printing service providers.
We did not include the cost of the PC, power supply unit, and emergency stop to the cost estimate.

\subsection{Sensitivity to External Disturbance}

The aim of this assessment is to show the sensitivity of the prototypes.
As an example, we use the CTCR prototype as it can be considered the most challenging among the prototypes.
A disturbance at the distal end of CTCR prototype's tube should be barely detectable, due to the fact that each tube is actuated at the other end of the tube's distal end and the applied flow and effort of the actuation module is not redirected to the distal end, unlike in a TDCR prototype.
Furthermore, the nested tubes are flexible and can store elastic energy distorting the transparency of the system.

\begin{figure}[!h]
    \centering
    \includegraphics[width=0.95\columnwidth]{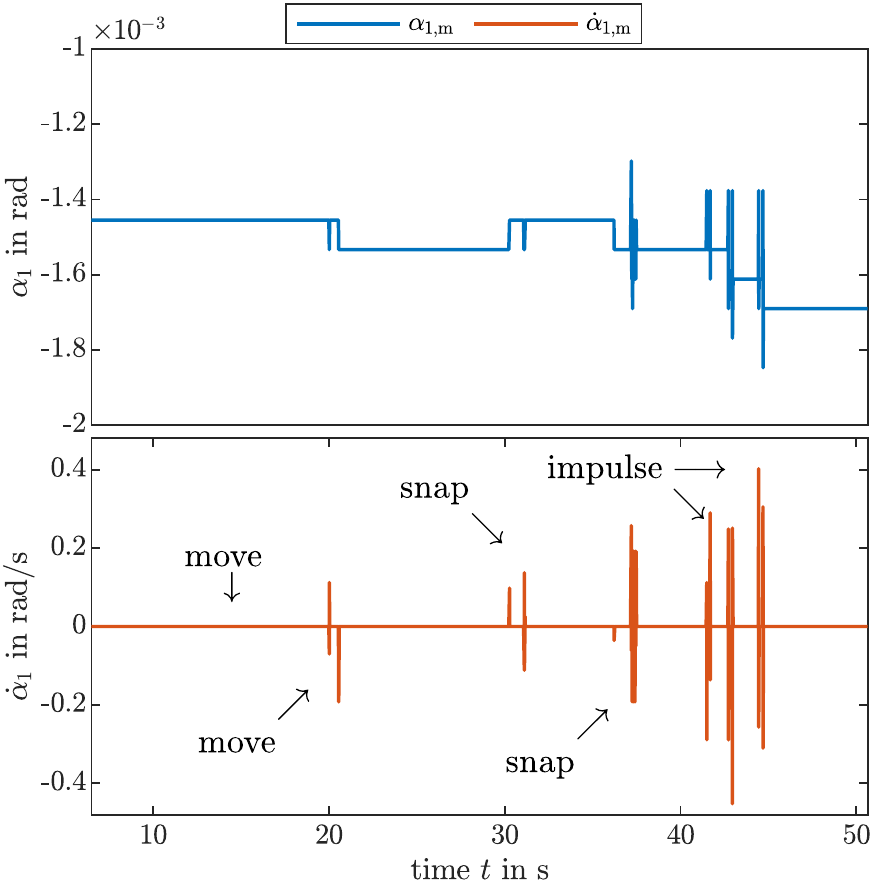}
    \caption{
    Proprioceptive measurement of external disturbances.
    The actuation module of the CTCR prototype is sensitive enough to different types of disturbances in the encoder level.
    See the accompanying video for a side-by-side comparison of the applied external disturbance and its measurements.
    }
    \label{fig:ctcr_snapping}
\end{figure}

For the assessment, an external disturbance is manually applied at the distal end of the most inner tube, while all actuators are in rest.
Figure~\ref{fig:ctcr_snapping} shows the detected values for the angle and angular velocity related to the rotational joint.
The value indicating the rotation of the outer tube is denoted by $\alpha_1$, where we use the notation described in \cite{GrassmannBurgner-Kahrs_et_al_IROS_2022}. 
Its time derivative is $\dot{\alpha}_1$ and an additional subscript as in $\dot{\alpha}_{1, \mathrm{m}}$ refers to a measured quantity.
Due to the backdrivability of the actuation module and the high-resolution encoder, the rotational joint angle $\alpha_1$ of the CTCR prototype can barely pick up on the different external disturbances.
Note that the corresponding motor current is too noisy for a conclusive detection.
However, the measured angular velocity $\dot{\alpha}_1$ shows a relation to external disturbances.
This shows that an external disturbance can be detected.

\section{DISCUSSION AND FUTURE WORK}

Our proposed actuation module allows building a variety of continuum robots and has the potential to serve as a common platform in the research community. 
Taking inspiration from hardware and electronics used in the Open Dynamic Robot Initiative, we created an actuation module with properties such as high-torque actuation, proprioceptive torque sensing, fast low-level motor control (\SI{10}{kHz}) and high-level control (\SI{1}{kHz}) with a fairly simple design at relatively low cost.

Evaluations on the toy examples show that tendon tension can be applied and measured by the actuation module.
The actuator's position is regulated, while gravity compensation for a known weight is considered as a feedforward term.
This can be seen as a proxy for tendon pre-tensioning.
Thanks to proprioception, this approach does not rely on additional sensors, e.g., tendon-tension sensors or load cells, reducing the implementation effort and cost.
This is thanks to the transparency and backdrivability of the actuation module, where the external load directly translates to motor torques, rather than being absorbed by a transmission with high gear-ratio.
The motor torque is also proportional to the motor current, which can be measured and, consequently, related to the external load.
Furthermore, evaluations on the toy example with the beam show an interesting direction for future work.
For instance, the influence of the beam bending is visible in the motor current raising the possibility to compensate it.
We leave this approach for future work, since it relies on an accurate dynamic model tailored to a specific continuum robot type.

The presented continuum robot prototypes inherit the capabilities of the actuation module.
Therefore, they are capable of torque control and fast trajectory tracking making them the first of a kind of the respective continuum robot type.
We show the potential of proprioceptive sensing with the CTCR prototype, which picks up on the external disturbances.
The difference of the motor position indicates a visible signal, while the motor position and current are insensitive or too noisy.
This assessment indicates the capability for interaction with the environment due to the ability to detect external disturbances in a proprioceptive manner.
Estimating the force's magnitude and direction with or without model-based approaches such as \cite{AloiRucker_et_al_RA-L_2022} is an important direction towards physical interactions.
Related to involuntary self-motion, the detection of snapping via motor velocity might be another promising future direction.
For instance, avoidance of snapping by utilizing an energy-based approach \cite{GilberthendrickWebster_T-RO_2015} with the proposed actuation module.

The presented actuation module as well as built prototypes are not free of limitations.
The actuation module can exert high torque, which might be necessary to apply for a specific prototype.
While the brushless motor is theoretically capable of producing very high motor torque, it is limited by electronics and power supply.
Note that this current will cause the motor winding to heat up and cooling is necessary, if applied for a longer duration.
Furthermore, gear slipping can occur due to dimensional tolerances, which causes higher friction between the gears.
Both slipping and friction can be addressed by improving the 3D printed parts.
In general, design improvements are inevitable and
will be made available$^\dagger$ in the future and updated regularly.

\section{CONCLUSIONS}

We propose an open-source actuation module that can measure tendon tension and joint torque in a proprioceptive manner.
The versatile and modular characteristics of the actuation module are illustrated by building three different types of continuum robot prototypes.

As we are observing a technological trend towards torque controlled robot systems in other research fields, we forecast a likely impact of the presented actuation module and continuum robot prototypes built on it.
While our assessments are of preliminary nature, they are an important foundation and a prerequisite towards advanced control methodologies.
We are confident that the ability of continuum robots utilizing our actuation module or one with similar capabilities, e.g., direct drive, have a decisive advantage over previous existing prototypes.

With our open actuation module, we aim to increase the reproducibility and accessibility of continuum robotics research by providing the hardware (CAD, parts, electronics components) alongside assembly instructions as well as software with our \href{http://www.opencontinuumrobotics.ca}{Open Continuum Robotics Project}$\dagger$.

\addcontentsline{toc}{section}{REFERENCES}
\bibliographystyle{IEEEtran}
\bibliography{literature}

\end{document}